\documentclass[journal]{IEEEtran}
\usepackage{amssymb, amsmath}
\usepackage{graphicx}
\usepackage{cleveref}
\usepackage{todonotes}
\usepackage{booktabs}
\usepackage{multirow}
\usepackage{verbatim}
\usepackage{url}
\usepackage{subcaption}
\ifCLASSINFOpdf
\else
\fi
\begin{document}
%
\title{Introducing the Hearthstone-AI Competition}
%
%
%
	\author{\IEEEauthorblockN{Alexander~Dockhorn and Sanaz~Mostaghim}\\
	\IEEEauthorblockA{Institute for Intelligent Cooperating Systems \\
	Department for Computer Science, Otto von Guericke University Magdeburg\\
	Email: \{alexander.dockhorn, sanaz.mostaghim\}@ovgu.de	}}

\maketitle

\begin{abstract}
The Hearthstone AI framework and competition motivates the development of artificial intelligence agents that can play collectible card games. A special feature of those games is the high variety of cards, which can be chosen by the players to create their own decks. In contrast to simpler card games, the value of many cards is determined by their possible synergies. The vast amount of possible decks, the randomness of the game, as well as the restricted information during the player's turn offer quite a hard challenge for the development of game-playing agents. This short paper introduces the competition framework and goes into more detail on the problems and challenges that need to be faced during the development process. 
\end{abstract}


%
\IEEEpeerreviewmaketitle

%
%
%
%

\section{Introduction}

\IEEEPARstart{T}{he}
development of artificial intelligence (AI) was often guided by the plethora of available real world applications. 
The recent success of AI agents such as AlphaGo, brought game AI back into the center of attention for media and researchers alike.
Games pose challenging and often well-balanced problems demanding the development of new agent architectures and allowing their evaluation based on their success in playing the game.
Due to their competitive nature, games offer an ideal test-bed for the comparison of multiple agents under differing conditions.

Game based competitions and benchmarks motivated researchers to create specialized agents in games of many different genres, such as Chess \cite{Campbell2002}, Go \cite{Silver2016}, Poker \cite{Brown2017}, Pac-Man \cite{Williams2016}, Starcraft \cite{Synnaeve2011} and many more.
While these focused on the development of an agent for a single game, other competitions and frameworks tried to generalize the solutions to a broader scope. 
The Arcade Learning Environment framework (ALE) \cite{Bellemare2013} as well as the General Video Game AI framework (GVGAI) \cite{Perez-Liebana2016} test an agent's success on a wide range of games.
All of these benchmarks pose unique demands on the agent's planning and reasoning capabilities, while there accessibility ensures a simple generation and evaluation of new agents.

In this paper we introduce the Hearthstone AI competition, because we believe that it is an excellent addition to the set of currently available benchmarks.
The focus of this competition is the development of autonomously playing agents in the context of the online collectible card game Hearthstone.

\newpage
Interesting features of collectible card games include, but are not limited to:
\begin{itemize}
    \item \textbf{Partial observable state space:} Critical information is hidden from the player. The agent typically does not know which cards it will draw and is unaware of the opponent's deck and hand cards. This is especially relevant when estimating the risk of an action, since this often depends on the current options of our opponent.
    \item \textbf{High complexity:} Hearthstone currently features more than 2000 different cards. This high amount and the number of their unique effects drastically increases the game-tree complexity.
	\item \textbf{Randomness:} In contrast to similar deck-building card-games, card effects in Hearthstone often involve randomness. This makes it particularly difficult to plan ahead, such that the agent needs to continuously adapt its strategy according to the observed result.
	\item \textbf{Deck-building:} A deck in Hearthstone consists of 30 cards, which do not necessarily need to be unique. 
	Card synergies mean that certain combinations of cards often have a stronger effect than playing the respective cards separately.
	Exploiting these synergies is a very complex task but allows to play cards to their full potential.
	\item \textbf{Dynamic Meta-Game:}
	The odds of winning the next game depend not only on the player's skill or the current deck, but also on the rate of other decks being played.
	The analysis of this meta-game can be crucial for creating or choosing the next deck to be played. 
	Finally, a deep understanding of the meta-game can be used to predict future enemy moves based on previously seen cards and select appropriate actions based on them.
\end{itemize}

In the following  we will give a short introduction to Hearthstone (\Cref{hearthstone}) and present the competition framework (\Cref{competition}).
Competition tracks of the Hearthstone AI'19 competition are highlighted in  after which an outline of future competition tracks is presented (\Cref{competition-tracks}).

More information on the competition, results of the 2018's installment and additional resources for research on Hearthstone can be found at:\\
\hphantom{.......}  \url{http://www.ci.ovgu.de/Research/HearthstoneAI.html}

We invite all developers to participate in this exciting research topic and hope to receive many interesting submissions.

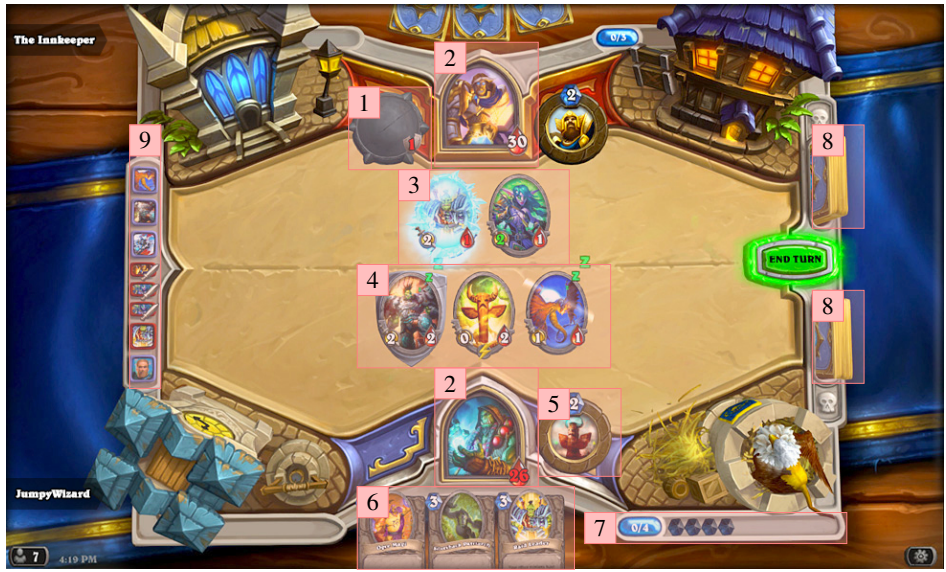
\begin{figure*}[t]
	\centering
	\resizebox{0.7\textwidth}{!}{
    	\begin{tikzpicture}
    	\node[anchor=south west,inner sep=0] at (0,0) {\includegraphics[width=0.75\textwidth]{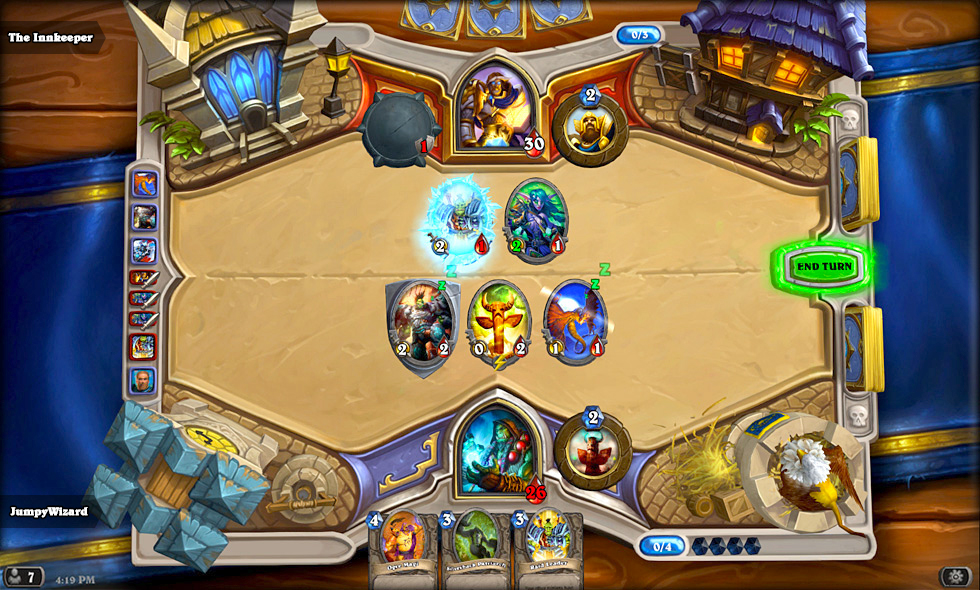}};
    	\draw[red!50,fill=red!25,fill opacity=0.3] (4.975,5.8) rectangle (6.175,7.0);	
    	\draw[red!50,fill=red!25] (4.975,7.0) rectangle (5.425,6.55) node[pos=.5, black] { 1};
    	
    	\draw[red!50,fill=red!25,fill opacity=0.3] (6.225,5.85) rectangle (7.725,7.65);	
    	\draw[red!50,fill=red!25] (6.225,7.2) rectangle (6.675,7.65) node[pos=.5, black] { 2};
    	
    	\draw[red!50,fill=red!25,fill opacity=0.3] (6.225,1.2) rectangle (7.725,2.925);	
    	\draw[red!50,fill=red!25] (6.225,2.925) rectangle (6.675,2.475) node[pos=.5, black] { 2};
    	
    	\draw[red!50,fill=red!25,fill opacity=0.3] (5.7,4.425) rectangle (8.175,5.8);	
    	\draw[red!50,fill=red!25] (5.7,5.8) rectangle (6.15,5.35) node[pos=.5, black] { 3};
    	
    	\draw[red!50,fill=red!25,fill opacity=0.3] (5.1,2.925) rectangle (8.775,4.425);	
    	\draw[red!50,fill=red!25] (5.1,4.425) rectangle (5.55,3.975) node[pos=.5, black] { 4};

    	\draw[red!50,fill=red!25,fill opacity=0.3] (7.725,1.35) rectangle (8.925,2.625);	
    	\draw[red!50,fill=red!25] (7.725,2.625) rectangle (8.175,2.175) node[pos=.5, black] { 5};
    	
    	\draw[red!50,fill=red!25,fill opacity=0.3] (5.1,0.0) rectangle (8.25,1.2);	
    	\draw[red!50,fill=red!25] (5.1,1.2) rectangle (5.55,0.75) node[pos=.5, black] { 6};
    	
    	\draw[red!50,fill=red!25,fill opacity=0.3] (8.4,0.375) rectangle (12.2,0.825);	
    	\draw[red!50,fill=red!25] (8.4,0.825) rectangle (8.85,0.375) node[pos=.5, black] { 7};
    	
    	\draw[red!50,fill=red!25,fill opacity=0.3] (11.7,2.7) rectangle (12.45,4.05);	
    	\draw[red!50,fill=red!25] (11.7,4.05) rectangle (12.15,3.6) node[pos=.5, black] { 8};
    	\draw[red!50,fill=red!25,fill opacity=0.3] (11.7,4.95) rectangle (12.45,6.45);	
    	\draw[red!50,fill=red!25] (11.7,6.45) rectangle (12.15,6) node[pos=.5, black] { 8};
    	
    	\draw[red!50,fill=red!25,fill opacity=0.3] (1.8,2.625) rectangle (2.25,6.45);	
    	\draw[red!50,fill=red!25] (1.8,6.45) rectangle (2.25,6) node[pos=.5, black] { 9};

    	\end{tikzpicture}
    }
	\caption{Elements of the Hearthstone game board: (1) weapon slot (2) hero (bottom: player, top: opponent) (3) opponent's minions, (4) player's minions, (5) hero power, (6) hand cards, (7) mana, (8) decks, (9) history}
	\label{fig:game-board}
\end{figure*}	

\newpage
\section{Hearthstone: Heroes of Warcraft}
\label{hearthstone}
Hearthstone is a turn-based digital collectible card game developed and published by Blizzard Entertainment \cite{hearthstone}.
Players compete in one versus one duels using self-constructed decks belonging to one hero out of nine available hero-classes.
In those matches players try to beat their opponents by reducing their starting health from 30 to 0.
This can be achieved by playing cards from the hand onto the game board at the cost of \textit{mana}. 
Played cards can be used to inflict damage to the opponent's hero or to destroy cards on his side of the game board.
The amount of mana available to the player increases every turn (up to a maximum of 10).
More mana gives access to increasingly powerful cards and increases the complexity of turn while the game progresses.
At the beginning of each turn the player draws a new card until his deck is empty, in which case he receives a step-wise increasing amount of \textit{fatigue}-damage.
The standard game board is shown in \Cref{fig:game-board}.

Players need to construct their own decks to play the game.
Those consist of 30 cards, which can be chosen out of more than 2000 currently available cards.
Cards and game mechanics are added in regular updates.
Each card bears unique effects, which the players can use to their advantage.
Additionally, each player chooses a hero, which gives access to a class specific pool of cards and hero power.

\begin{figure*}[t]
	\begin{subfigure}{0.3\textwidth}
		\begin{tikzpicture}
		\node[anchor=south west,inner sep=0] at (0,0) {\includegraphics[width=0.9\textwidth]{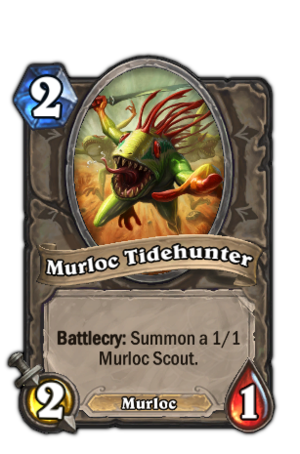}};
		
		\draw[red!50,fill=red!25,fill opacity=0.3] (.15,5.325) rectangle (1.35,6.525);	
		\draw[red!50,fill=red!25] (0.15,6.075) rectangle (0.6,6.525) node[pos=.5, black] { 1};
		
		\draw[red!50,fill=red!25,fill opacity=0.3] (.2,0.2) rectangle (1.35,1.35);	
		\draw[red!50,fill=red!25] (0.2,0.9) rectangle (0.6,1.35) node[pos=.5, black] { 2};
		
		\draw[red!50,fill=red!25,fill opacity=0.3] (3.475,0.2) rectangle (4.625,1.35);	
		\draw[red!50,fill=red!25] (3.475,0.9) rectangle (3.925,1.35) node[pos=.5, black] { 3};
		
		\draw[red!50,fill=red!25,fill opacity=0.3] (.4,1.35) rectangle (4.425,2.25);	
		\draw[red!50,fill=red!25] (0.4,2.25) rectangle (0.8,1.8) node[pos=.5, black] { 4};
		\end{tikzpicture}
		\centering
		\caption{Minion card}
	\end{subfigure}
	\hfill
	\begin{subfigure}{0.3\textwidth}
		\centering
		\begin{tikzpicture}
		\node[anchor=south west,inner sep=0] at (0,0) {\includegraphics[width=0.9\textwidth]{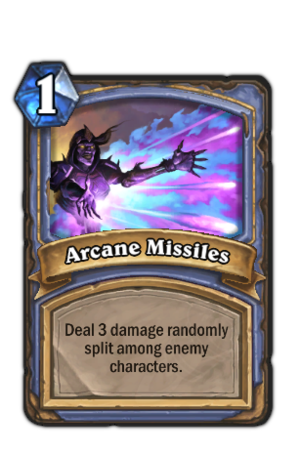}};
		
		\draw[red!50,fill=red!25,fill opacity=0.3] (.15,5.325) rectangle (1.35,6.525);	
		\draw[red!50,fill=red!25] (0.15,6.075) rectangle (0.6,6.525) node[pos=.5, black] { 1};
		
		\draw[red!50,fill=red!25,fill opacity=0.3] (.4,1.15) rectangle (4.425,2.25);	
		\draw[red!50,fill=red!25] (0.4,2.25) rectangle (0.8,1.8) node[pos=.5, black] { 4};
		\end{tikzpicture}
		
		\caption{Spell card}
	\end{subfigure}
	\hfill
	\begin{subfigure}{0.3\textwidth}
		\centering
		\begin{tikzpicture}
		\node[anchor=south west,inner sep=0] at (0,0) {\includegraphics[width=0.91\textwidth]{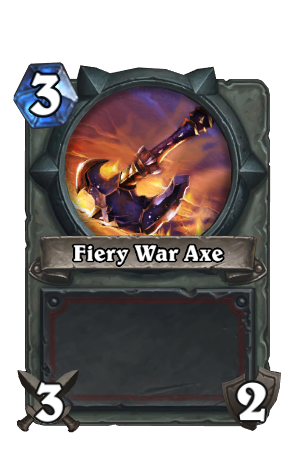}};
		
		\draw[red!50,fill=red!25,fill opacity=0.3] (.15,5.325) rectangle (1.35,6.525);	
		\draw[red!50,fill=red!25] (0.15,6.075) rectangle (0.6,6.525) node[pos=.5, black] { 1};
		
		\draw[red!50,fill=red!25,fill opacity=0.3] (.2,0.2) rectangle (1.35,1.35);	
		\draw[red!50,fill=red!25] (0.2,0.9) rectangle (0.6,1.35) node[pos=.5, black] { 2};
		
		\draw[red!50,fill=red!25,fill opacity=0.3] (3.475,0.2) rectangle (4.625,1.35);	
		\draw[red!50,fill=red!25] (3.475,0.9) rectangle (3.925,1.35) node[pos=.5, black] { 3};
		\end{tikzpicture}
		
		\caption{Weapon card}
	\end{subfigure}
	\caption{General card types: cards include (1) mana cost, (2) attack damage, \mbox{(3) health/durability}, \mbox{(4) and special effects}.}
	\label{fig:card_types}
\end{figure*}

Cards can be of the type minion, spell, or weapon.
\Cref{fig:card_types} shows one example of each card type.
Minion cards assist and fight on behalf of the hero. 
They usually have an attack, health, and mana cost-value, as well as a short ability text.
Furthermore, minions can belong to a special minion type, which is the basis for many synergy effects.
Once played, they can attack the enemies side of the board in every consecutive turn to inflict damage on either the opponent's minions or hero.
Attacking a target also reduces the attacker's health by the target's attack value.
In case any minion's health drops to zero, it is removed from the board and put into its player's graveyard.
Spell cards can be cast at the cost of mana to activate various abilities and are discarded after use.
They can have a wide range of effects, e.g. raising a minion's attack or inflicting damage to a random minion.
Secrets, which are a special kind of spells, can be played without immediately activating their effect. 
After a trigger condition was fulfilled, the secret will be activated. 
Once activated, the secret is removed from the board.
Weapon cards are directly equipped to the player's hero and enable him to attack.
Their durability value limits the number of attacks till the weapon breaks.
Only one weapon can be equipped at the same time.

Hearthstone decks are often created around a common theme.
Multiple cards that positively influence each other can create strong synergies and increase the value of each card in context of its deck.
For this reason, the value of a single card highly depends on the player's hand, current elements on the board, and the deck in general.
Common examples are minion cards of the same type, e.g. ''Murloc``, which give each other additional advantages, e.g. an attack boost.
Each of these minions is comparatively weak, but their value increases when they are played together.

Generated decks can be categorized into three major categories: aggro, mid-range, and control.
Aggro decks build on purely offensive strategies, which often include a lot of minions.
Control decks try to win in the long run by preventing the opponent's strategy and dominating the game situation.
The playing style of mid-range decks is between aggro and control.
They try to counter early attacks to dominate the game board with high-cost minions in the middle of the game.

Game length and branching factor can be dependent on the player's decks in the current game.
Some decks try to play single high-cost cards, whereas others build on versatile combinations.
The complexity of each turn and the uncertainty faced during the game makes Hearthstone a challenging problem for AI research.

\section{Hearthstone-AI Competition Framework}
\label{competition}



The Hearthstone-AI competition is based on the community driven simulator Sabberstone.
This framework is written in C\# and the competition extends the original framework by multiple helper classes which provide an agent simple means of accessing the current game state limited to variables that would have been observable to a human player.

More specifically, each agent needs to inherit from the \emph{AbstractAgent} class.
The included functions \emph{InitializeAgent} and \emph{FinalizeAgent} can be used to load and store information at the beginning and end of each session.
Additionally, \emph{InitializeGame} and \emph{FinalizeGame} are called at the beginning and end of each simulated game.
These functions can be used to setup a strategy or updating it based on the games' outcome.

During a game each time the agent needs to choose an action its \emph{GetMove} function is called.
The agent is given a \emph{POGame} object representing the partial observation of the current game-state. It contains information about the visible part of the game board, a set of remaining cards in its deck, its hand cards, and the number of cards in their opponents' hand.
Furthermore, the opponent's deck as well as its hand cards are replaced by dummy-cards to ensure that this information remains hidden to the agent. 
The agent is ensured 60s of computation time to step-wise return a set of actions and concluding its turn.
In case the turn was not ended by the agent, the returned action will be processed irreversibly and an updated game-state will be returned to the agent while asking for its next action.
Actions of both players are applied until a winner can be determined or a maximum number of turns (default = 50) is exceeded. In the latter the game ends with a draw.

The \emph{POGameHandler} class controls the simulation of multiple games and reports the result of these simulations in terms of a \emph{GameStats} object. The number of wins, draws, and loses as well as the total and average response times per agent are tracked and reported at the end of a simulation session.
\section{Competition Tracks}
\label{competition-tracks}

During the first years of the Hearthstone-AI competition two tracks will be open for entry. We plan to extend this list in the following years to give users some time to accommodate with the framework and the game itself. 
The following two tracks will be open for submission in the Hearthstone-AI'19:
\begin{itemize}
	\item \textbf{Premade Deck Playing”-track}
	In the “Premade Deck Playing”-track participants will receive a list of six decks and play out all combinations against each other. Only three of the six decks will be known to the developers before the final submission.
	Determining and using the characteristics of player’s and the opponent’s deck to the player’s advantage will help in winning the game.
	The long-term goal of this track will be the development of an agent that is capable of playing any deck.
	
	\item \textbf{User Created Deck Playing-track}
	The “User Created Deck Playing”-track invites all participants to create their own decks or to choose from the vast amount of decks available online. 
	Finding combinations of decks and agents that can consistently beat others will play a key role in this competition track. 
	Additionally, it gives the participants the chance to optimize the agent’s strategy to the characteristics of their chosen deck.
\end{itemize}

A round robin tournament will be used to determine the average win-rate of each agent and rank them accordingly.
In case this process becomes unfeasible due to a large number of participants, we will split submissions into multiple sub-tournaments to determine the best performing agents among them and use a round robin tournament to determine the winner of the competition.
Matches will be repeated multiple times to accommodate for the randomness in the card draw.

In order to support incremental improvements of the agents' performance, we plan to make all submissions publicly available on the competition website after the competition evaluation has been completed.

\newpage

\section{Conclusions and Future Plans}
\label{conclusion}

Previous work on collectible card games has been scattered on different games and frameworks.
With this competition we want to provide a unified way to develop and compare AI approaches on multiple collectible card game related tasks.
While current competition tracks focus on the agents' basic game playing capabilties, we like to cover various different tasks in future installments of this competition.
Specifically, we plan to implement the following future tracks:
\begin{itemize}
	\item \textbf{Deck-building:} As soon as agents are reasonably skilled in playing a given deck we want to further explore the deck building process. This is currently considered a highly creative task in which many game characteristics need to be considered to build outstanding decks.
	\item \textbf{Draft mode deck-building:} A special form of deck building is the draft mode. Here, an agent is presented 3 cards at a time of which it needs to choose one to include it in its deck. This process is repeated until the deck is filled. Strategic planning and estimating the value of each card are important to exploit synergies and building a competitive deck.
	\item \textbf{Game balancing/Card generation:} During the development process of a collectible card game much of the time is put into the generation of new game mechanics and cards, since these need to fit and complement the current card pool. A future track will aim to explore these balancing tasks in more detail.
\end{itemize}
\vspace{1em}

We hope that this short introduction of our competition framework motivates further research in this very interesting topic.
More information on how to participate in the Hearthstone-AI'19 competition, results of the 2018's installment and additional resources for research on Hearthstone can be found at:\\
\hphantom{.......}  \url{http://www.ci.ovgu.de/Research/HearthstoneAI.html}

\vspace{1cm}
\section*{Acknowledgement}

We would like to thank all who contributed to the Sabberstone framework on which this competition is based on. Special thanks goes to \emph{darkfriend77} and \emph{Milva} who are currently organizing the framework's ongoing development process. This competition would not have been possible without them.

\newpage

\ifCLASSOPTIONcaptionsoff
  \newpage
\fi



%
\bibliographystyle{bib/IEEEtran}

\bibliography{ms}

%

\vspace{-6.0cm}
\begin{IEEEbiography}[{\includegraphics[width=1in,height=1.25in,clip,keepaspectratio]{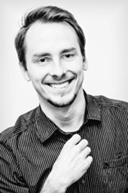}}]{Alexander Dockhorn} is PhD student at the Otto von Guericke University
in Magdeburg, Germany. In his research, he combines computational intelligence
in games and intelligent data analysis with focus on partial
information games and the active learning of rules and strategies from
successful plays. He is student member of the IEEE and its
Computational Intelligence Society. Currently he is vice-chair of the Student
Games-Based Competitions Sub-Committee.
\end{IEEEbiography}

\vspace{-6.05cm}
\begin{IEEEbiography}[{\includegraphics[width=1in,height=1.25in,clip,keepaspectratio]{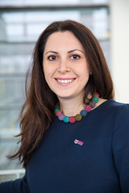}}]{Sanaz Mostaghim} is a professor of computer science at the Otto von
Guericke University of Magdeburg, Germany. Sanaz has received a
PhD degree (2004) in electrical engineering from the University of
Paderborn in Germany and worked as a post-doctoral fellow (2004 -
2006) at the Swiss Federal Institute of Technology (ETH) Zurich in
Switzerland. Sanaz received her habilitation in applied computer
science from Karlsruhe Institute of technology (KIT) in 2012, where she
worked as a lecturer (2006 - 2013). Her research interests are
evolutionary multi-objective algorithms, swarm intelligence and their
applications in science and industry. She is an active member in several
German and international societies such as IEEE, ACM and GI.
Currently she is the chair of Task Force Evolutionary Multi-objective Optimization (TFEMO) and is serving as an associate editor for IEEE Transactions on Evolutionary
Computation, IEEE Transactions on Emerging Technologies and IEEE Transactions on
Cybernetics.
\end{IEEEbiography}





\end{document}